\def\paperTitle{
Solving Math Word Problems via Cooperative Reasoning induced Language Models
}

\newcommand*{\affmark}[1][*]{\textsuperscript{#1}}

\def\authorBlock{
    Xinyu Zhu\affmark[$\diamondsuit$]\footnotemark[1] \qquad
    Junjie Wang\affmark[$\spadesuit$]\footnotemark[1] \qquad
    Lin Zhang\affmark[$\heartsuit$] \qquad
    Yuxiang Zhang\affmark[$\spadesuit$]
    \\
    \textbf{
    Ruyi Gan\affmark[$\heartsuit$] \qquad
    Jiaxing Zhang\affmark[$\heartsuit$] \qquad
    Yujiu Yang\affmark[$\diamondsuit$]\footnotemark[2]
    } \\
    \affmark[$\diamondsuit$]Tsinghua University \quad
    \affmark[$\spadesuit$]Waseda University \quad \\
    \affmark[$\heartsuit$]International Digital Economy Academy
    \\
    {\tt\small zhuxy21@mails.tsinghua.edu.cn} \qquad {\tt\small yang.yujiu@sz.tsinghua.edu.cn} \\
    {\tt\small wjj1020181822@toki.waseda.jp} \qquad {\tt\small joel0495@asagi.waseda.jp} \\
    {\tt\small \{zhanglin, ganruyi, zhangjiaxing\}@idea.edu.cn}
}

\newif\ifreview 
\newif\ifarxiv 
\newif\ifcamera \newcommand{\cameraready}{\cameratrue}
\cameraready

\documentclass[11pt]{article}
\ifreview \usepackage[review]{ACL2023} \fi
\ifarxiv \usepackage{ACL2023} \fi
\ifcamera \usepackage{ACL2023} \fi

\usepackage{times}
\usepackage{latexsym}
\usepackage[T1]{fontenc}
\usepackage[utf8]{inputenc}
\usepackage{microtype}
\usepackage{inconsolata}

\usepackage{multirow}

\usepackage{graphicx}
\usepackage{amsmath}
\usepackage{amssymb}
\usepackage{booktabs}
\usepackage{amsfonts}
\usepackage{algorithm}
\usepackage{algorithmic}
\usepackage{footnote}

\usepackage{xcolor}
\usepackage{comment}
\usepackage{amsthm}
\usepackage{bbm} 
\usepackage{subcaption}
\usepackage{multirow}
\usepackage{comment}
\usepackage{makecell} 
\usepackage{tabularx}
\usepackage[normalem]{ulem}
\useunder{\uline}{\ul}{}
\usepackage{booktabs}
\usepackage[switch]{lineno}

\usepackage{url}
\usepackage{xspace}

\usepackage[export]{adjustbox}
\usepackage{paralist} 

\usepackage{tabu}

\newcommand{\nbf}[1]{{\noindent \textbf{#1}}}

\newcommand{\eat}[1]{}

\newcommand{\cbit}{\begin{compactitem}}
\newcommand{\ceit}{\end{compactitem}}
\newcommand{\cben}{\begin{compactenum}}
\newcommand{\ceen}{\end{compactenum}}

\usepackage[capitalize]{cleveref}
\crefname{section}{Sec.}{Secs.}
\crefname{table}{Table}{Tables}
\crefname{figure}{Fig.}{Figs.}

\title{\paperTitle}
\author{\authorBlock}

\begin{document}
\maketitle

{
  \renewcommand{\thefootnote}%
  {\fnsymbol{footnote}}
  \footnotetext[1]{Equal contribution.}
  \footnotetext[2]{Corresponding Author.}
}

\begin{abstract}

Large-scale pre-trained language models (PLMs) bring new opportunities to challenging problems, especially those that need high-level intelligence, such as the math word problem (MWPs).
However, directly applying existing PLMs to MWPs can fail as the generation process lacks sufficient supervision and thus lacks fast adaptivity as humans.
We notice that human reasoning has a dual reasoning framework that consists of an immediate reaction system (system 1) and a delicate reasoning system (system 2), where the entire reasoning is determined by their interaction.
This inspires us to develop a cooperative reasoning-induced PLM for solving MWPs, called \textbf{Co}operative \textbf{Re}asoning (\textbf{CoRe}), resulting in a human-like reasoning architecture with system 1 as the generator and system 2 as the verifier.
In our approach, the generator is responsible for generating reasoning paths, and the verifiers are used to supervise the evaluation in order to obtain reliable feedback for the generator.
We evaluate our CoRe framework on several mathematical reasoning datasets and achieve decent improvement over state-of-the-art methods, up to $9.6\%$ increase over best baselines.\footnote{Our codes are available at \url{https://github.com/TianHongZXY/CoRe}}

\end{abstract}

\section{Introduction}
\label{sec:introduction}

\begin{figure}[!tp]
    \centering
    \includegraphics[width=0.48\textwidth]{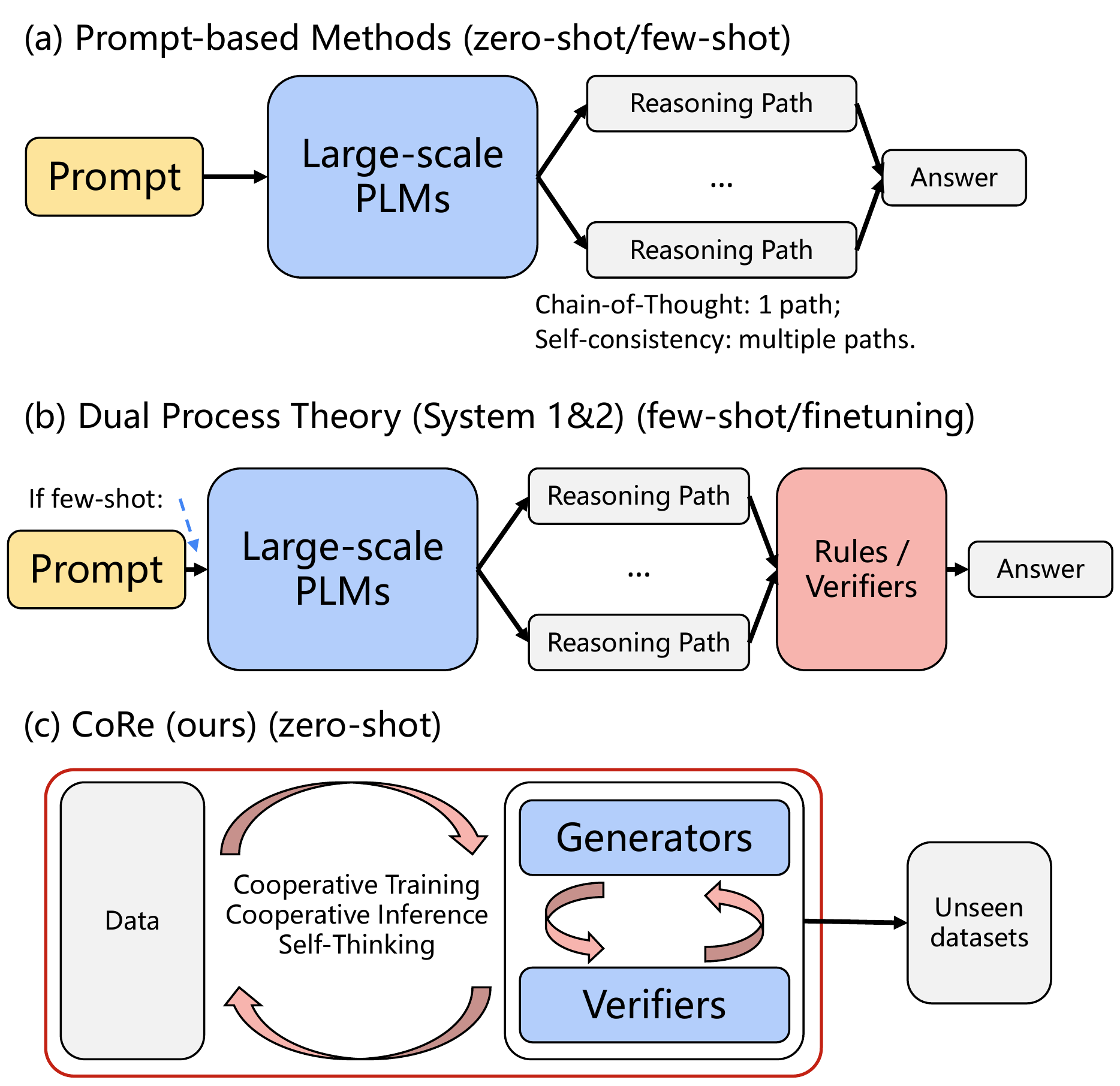}
    \caption{
    Comparing our CoRe with popular methods in mathematical logic reasoning tasks.
    }
    \label{fig:introduction}
\end{figure}

\eat
{
How can Artificial Intelligent (AI) learn to reason math problems as human does?
Furthermore, we expect it can be deployed with unseen math problems.
Fortunately, recent remarkable works provide several solutions from two perspectives. 
On the one hand, thanks to numerous parameters ($\geq$ 100B) and vast data, the large-scale Pre-trained Language Models (PLMs) earns emergence~\cite{DBLP:journals/corr/abs-2204-02311@palm, emergent}.
They can be guided to generate reasoning paths by well-designed trigger prompts for few-shot (zero-shot) reasoning tasks~\cite{DBLP:journals/corr/abs-2108-07258@foundation_models,DBLP:journals/corr/abs-2201-11903@cot-ori, DBLP:journals/corr/abs-2205-11916@zero-shot-reasoners}.
Furthermore, considering multiple reasoning paths and then voting also help PLMs~\cite{DBLP:journals/corr/abs-2203-11171@self-consistency, diverse-cot, DBLP:conf/acl/RajaniMXS19@explain-yourself}.
On the other hand, several works~\cite{DBLP:conf/nips/NyeTTL21@coherence-and-consistency,openai@gsm8k} draw insight from the dual process theory~\cite{daniel2017@thinking-fast-slow,EVANS2003454@in-two-minds} to reasoning field and achieve improvements.
In detail, they treat PLMs as System 1 (automatic, non-conscious, for generation) and verifiers as System 2 (controlled, conscious, for evaluation) of human. %
However, they are processed as two-step workflows by separating System 1\&2.
In contrast, we argue that \textit{an effective cooperative working with PLMs can implement a zero-shot math reasoner.}
}

Addressing math problems is a hallmark of human intelligence, which allows reasoning and adapting from limited data. 
We want neural models to be able to do the same,
however, quick and flexible reasoning is challenging to current neural models as they must possess a certain level of prior experience from a limited amount of new data while avoiding overfitting.
The rapid growth of large-scale Pre-trained Language Models (PLMs) offers unprecedented potential for this issue, often relying on well-designed trigger prompts~\citep{DBLP:journals/corr/abs-2201-11903@cot-ori, diverse-cot, DBLP:conf/nips/BrownMRSKDNSSAA20@gpt3}.
Although appealing in terms of efficiency, its success relies on memorizing patterns with a sufficiently large number of parameters  ($\geq$ 100 billion)~\cite{DBLP:journals/corr/abs-2206-07682@emergent}, differentiating it from the fast adaptivity in the human reasoning process.
Active disciplines like neuroscience and cognitive science attempt to uncover the mechanism of human reasoning, and agree that our learning process is governed by an interaction mechanism, often referred to as System 1 and System 2~\citep{EVANS2003454@in-two-minds, daniel2017@thinking-fast-slow}.
In particular, System 1 offers fast responses like human instinct, and System 2 performs deliberate reasoning.
Interactions between them are important for adapting to a continuously changing environment.
PLMs behave more like System 1, according to the above theory, and thus lack the generalization ability in reasoning~\citep{DBLP:conf/nips/NyeTTL21@coherence-and-consistency}.

In this work, we explore a new line of zero-shot math problem reasoning, using a human reasoning-alike framework with feedback in the solution generation loop as opposed to pure PLM-based methods, called \textbf{Co}operative \textbf{Re}asoning (\textbf{CoRe}).
Intuitively, System 1 and System 2 are embodied as generators and verifiers, respectively, and they are defined as follows: generators for generating reasoning paths, and verifiers for supervising the paths' evaluation.
Specifically, we train a LM beyond the question-answer paradigm by integrating in-the-loop reasoning, i.e., we let the LM output both the answer and the corresponding reasoning process for a given question.
Meanwhile, we introduce two types of verifiers, including token-level and sentence-level, allowing us to provide feedback in the whole solution generation lifecycle.
Notice that the solution path is generated by selecting candidate tokens with some probability so that it is tree-alike and much coincides with the tree search process of Monte Carlo Tree Search (MCTS)~\cite{DBLP:conf/ecml/KocsisS06@mcts}.
With this in mind, the verifiers can score tokens along the solution generation process from start to end when using the MCTS.
Therefore, we can use the score to evaluate the quality of the generation process during inferring before finalizing the solution, making timely feedback available for supervising the generation process.
With this, the evaluation goes beyond the quality of the final result at the granularity of each reasoning step, extending the supervision from the solution level to the path level.
We combine the solution score and the perplexity of its corresponding reasoning path to encourage the overall training towards high-quality augmented solutions while aligning with the reliable reasoning process, aiming to improve generalization ability.
{
Our experimentally evaluate CoRe on multiple mathematical reasoning datasets in both zero-shot and {fine-tuning} settings.
CoRe consistently achieves better performance than competing baselines.
Notably, CoRe has up to $9.6\%$ improvements on MultiArith over SoTA baselines, which are dozens of times larger than our model.
}

{
In summary, our contributions are as follows.
\begin{itemize}
    \item We propose a novel reasoning method for mathematical problem solving, called \textbf{Co}operative \textbf{Re}asoning (\textbf{CoRe}), that introduces feedback in the loop during solution generation as opposed to the sequential learning process in the previous ones, resulting in the first method for this task that builds on top of the learning mechanism in the human brain.
    \item We develop a self-thinking strategy for further boosting reasoning ability with generated data from the cooperation between System 1 and System 2.
    \item We demonstrate the superiority of CoRe comparing to other zero-shot and fine-tuning methods, which has $9.6\%$ improvements on MultiArith over SoTA baselines.
\end{itemize}
}

\section{Related Work}
\label{sec:related_work}

\subsection{Dual Process System}
\label{Dual_Process_System}

%
Dual-process theory~\citep{EVANS2003454@in-two-minds, daniel2017@thinking-fast-slow} argues there are two cognitive systems underpinning human reasoning: System 1 and System 2. The purpose of clarifying these systems is that they have the potential to help us construct artificial intelligence systems that benefit from human flexibility and methodical generalization.

Dual process system model guidance is not new. \citet{DBLP:conf/nips/NyeTTL21@coherence-and-consistency} simulated Systems 1 and 2 to improve consistency and coherence of neural networks.
Similar to several studies~\citet{openai@gsm8k,diverse-cot, mcts-gans}, in addition to System 1 for the generation, we develop a distinct model as System 2, called Verifier. 
The Verifier checks the feasibility and correctness of the generator's content and collaboratively solves the reasoning task together.

\subsection{Multi-step Reasoning}
\label{Multi-step_Reasoning}

Many works exploit the multi-step reasoning ability of language models.
\citet{openai@gsm8k} showed that training a verifier to score the solutions generated by a fine-tuned GPT-3 could improve the performance compared to solely fine-tuning a GPT-3. 
\citet{scratchpads} discovered that asking the language model to write the intermediate process could achieve better results on various NLP tasks. 
Likewise, Chain-of-Thought (CoT) prompts~\citep{DBLP:journals/corr/abs-2201-11903@cot-ori} prepended exemplars with intermediate reasoning steps as prompts and achieved SoTA on several reasoning benchmarks by using large-scale PLMs. 
\citet{DBLP:journals/corr/abs-2203-11171@self-consistency}~further boosted CoT's performance by sampling a bunch of possible solutions and then obtained the final answer by majority voting. 
DIVERSE~\citep{diverse-cot} proved diverse CoT prompts and an extra verifier were both helpful for PLMs to solve reasoning problems. \citet{DBLP:journals/corr/abs-2205-11916@zero-shot-reasoners}~found that by simply adding ``Let's think step by step'' after the question. PLMs could successfully step by step solve the problems, called Zero-shot-CoT.

These above methods rely on extremely large language models, resulting in high computational cost and time-consuming. 
Moreover, several works~\citep{DBLP:journals/corr/abs-2201-11903@cot-ori,DBLP:journals/corr/abs-2205-11916@zero-shot-reasoners} point out that neither CoT nor Zero-shot-CoT is helpful to smaller models. 
While our method does not necessarily require extremely large PLMs and can work with models with different size scales, thus reducing computational cost and inference time. 
Our approach has competitive zero-shot performance thanks to the efficient and collaborative application of a dual-process system.

\section{Cooperative Reasoning}
\label{sec:method}

\begin{figure*}[!tp]
    \centering
    \includegraphics[width=0.98\textwidth]{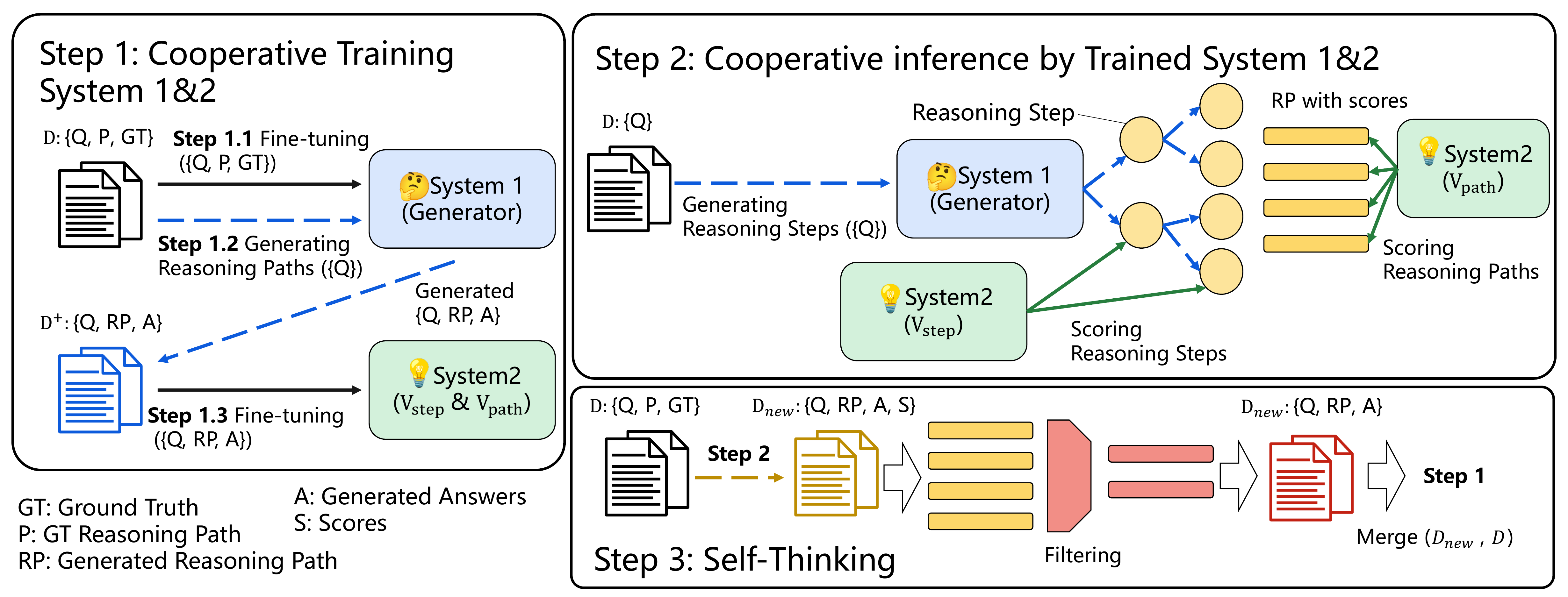}
    \caption{
    Cooperative reasoning framework.
    }
    \label{fig:whole_model}
\end{figure*} 

In this section, we will present the proposed cooperative reasoning framework, CoRe, that enforces System 1 and System 2 mutually cooperating, which includes $3$ sequential steps: cooperative training, cooperative inference, and self-thinking.

\subsection{Preparation}
\label{sec:preparation}

As discussed in \cref{sec:introduction}, we expect a PLM ($G$) to fast generate multiple reasoning paths like System 1.
Then, considering that System 2 is responsible for deliberate evaluations of the reasoning paths, we employ two modules: a step verifier ($V_{step}$) for reasoning steps, and a path verifier ($V_{path}$) for reasoning paths.

\subsection{Cooperative Training}
\label{sec:cooperative_training}

Before applying System 1\&2 to inference, a critical issue for them is \textit{learn how to generate reasoning paths and evaluate reasoning steps/paths}. 
Inspired by a widely-used training strategy for reasoners~\cite{openai@gsm8k}, we present a cooperative training method as shown in~\cref{fig:whole_model} Step 1.
Moreover, we discuss hyper-parameter configurations and extra training details in~\cref{append:hyperparameters} and~\cref{append:details_training_verifiers}.

\nbf{Step 1.1:} We first fine-tune $G$ on a dataset $D=\left\{\left(q_{i}, p_{i}, gt_{i}\right)\right\}_{i=1}^{N}$ consisting of $N$ samples. 
Each sample $x$ is composed of a question $q$, a reasoning path $p$ and a ground truth answer $gt$. 
We fine-tuen $G$ with standard language modeling objective $\mathcal{L}_{LM}$ as~\cref{eq:lm_loss}.
\begin{equation}
\small
\mathcal{L}_{LM}=-\sum_{i=1}^{|p| + |gt|} \log P\left(x_{i} \mid x_{<i} \right)
\label{eq:lm_loss}
\end{equation}

\nbf{Step 1.2:} Once $G$ has learned how to generate solutions, we employ it on questions $q$ from $D$.
As a result, we obtain a new dataset $D^{+}=\left\{\left(q_{i},{rp}_{i,j}, a_{i,j}\right)\right\}^{i=1,\ldots,N}_{j=1,\ldots,M}$ with $M$ generated reasoning paths ($rp$) and answers ($a$) for each $q$. 

\nbf{Step 1.3:} Different from the popular methods, we train two verifiers to model human reasoning procedure with deliberate analysis for each step and the whole path.
To evaluate several reasoning steps in a path, we desire a token-level scorer, which is named step verifier $V_{step}$.
Therefore, we fine-tune a PLM with two tasks jointly: 1) the language modeling task mentioned before; 2) the verification task to predict a score for each token in the solution. The verification loss $\mathcal{L}_{VS}$ is calculated as the Mean Squared Error (MSE) of the predicted score with respect to the label as follows:
\begin{equation}
\small
\mathcal{L}_{VS} = \sum_{i=1}^{|rp|+|a|}(score_{i} - \mathbb{I}(a==gt))^2,
\label{eq:verifier_step_mse}
\end{equation}
where, $(rp, a)$ from $D^{+}$ and $gt$ with same $q$ from $D$.

On the other hand, we need a path-level scorer for reasoning paths.
Different from step verifier, 
we simply extract an overall presentation of the reasoning path for prediction. %
Specifically, we employ a BERT-like model and take the {\tt [CLS]} token to calculate MSE loss $\mathcal{L}_{VP}$ similar to $\mathcal{L}_{VS}$.

In summary, the overall training objective for verifiers is given by:
\begin{equation}
\small
\mathcal{L}_{V} = \mathcal{L}_{VS} + \mathcal{L}_{LM} + \mathcal{L}_{VP}.
\label{eq:verifier_overall_loss}
\end{equation}

\subsection{Cooperative Inference}
\label{sec:cooperative_inference}

After obtaining a generator and two verifiers, we propose cooperative inference to generate solutions for unseen questions.
Instead of treating verifiers as voters, we argue that verifiers should offer appropriate guidance and feedback during the reasoning process.
Therefore, we integrate a cooperative search algorithm.
In particular, we adopt the popular Monte Carlo Tree Search (MCTS)~\cite{DBLP:conf/ecml/KocsisS06@mcts} to enable controlled reasoning.
The cooperative inference starts from the root node, which preserves question tokens. We detail the cooperative inference process as follows.

\nbf{Selection.} If the current node has children, with $50\%$ probability, we select a node from its children with the modified PUCT formula~\cite{DBLP:conf/aips/CzechKK21} as~\cref{eq:selection}, %
\begin{equation}
\small
n^\ast =\mathop{\arg\max}\limits_{n \in \mathcal{C}} (R(n) + c_{puct}\pi(n|s) \frac{\sqrt{\sum_{b \in \mathcal{C}} N(s, b)}}{1+N(s, n)}),
\label{eq:selection}    
\end{equation}
where the state $s$ represents the sequence consisting of all tokens in the current search path. 
And, $N(s, n)$ means the times that node $n$ has been selected in state $s$. Reward $R(n)$ records all the scores received from the backup.
We perform selection again with the selected node as the current node.
Otherwise, we perform expansion once and choose the returned new node as current node.

\nbf{Expansion.} During expansion, the generator is required to generate a sequence of tokens based on the current state.
A new node is created to store the generated tokens and added to the current node's children. %
Then, $V_{step}$ evaluates the current reasoning path and predict a score ${score}_{step}$. %
Finally, the new node is returned.

\nbf{Roll-Out.} After selection and expansion, we start from the current node and let the generator complete the reasoning path until it meets {\tt [EOS]} token or reaches the max token length limit. Next, $V_{path}$ evaluates the whole reasoning path and produces a score ${score}_{path}$.
Remember that $V_{step}$ also provides a score ${score}_{step}$ during the expansion.
Therefore to leverage both scores, we introduce a hyper-parameter $\alpha$ to adjust their contributions to the node's reward, 
\begin{equation}
\small
s = {score}_{path} + \alpha \times {score}_{step}
\label{eq:score}
\end{equation}
where $s$ is the final score that each node receives by the backup.

\nbf{Backup.} We update the rewards back from the current node to the root node. 
The scores produced by verifiers are added to $R(n)$ and the visited time $N(s, n)$ is increased by $1$. 

\subsection{Self-Thinking}
\label{sec:self_thinking}

\begin{algorithm}[tp]
	\renewcommand{\algorithmicrequire}{\textbf{Input:}}
	\renewcommand{\algorithmicensure}{\textbf{Output:}}
	\caption{Self-Thinking}
	\label{alg:self_thinking}
	\begin{algorithmic}[1]
		\REQUIRE{Generator $G$; Step verifier $V_{step}$; Path verifier $V_{path}$}; Dataset $D$.
		\STATE Combine generator and verifiers with a cooperative search algorithm.
		\REPEAT
		\STATE Generate a new dataset $D_{new}$ from input questions.
		\STATE Filter $D_{new}$.
		\STATE Merge $D_{new}$ with $D$ in Step 1.
		\STATE Do Step 1.
		\STATE Do Step 2.
		\UNTIL{performance is saturated.}
	\end{algorithmic}  
\end{algorithm}

It is challenging to fine-tune models on the data synthesized by themselves, which indicates they have to be very confident in the content they generate.
A proper self-training method can enhance the robustness of the whole system and allow deep data mining.
Therefore, we introduce self-thinking as described in~\cref{fig:whole_model} Step 3 and~\cref{alg:self_thinking}.
Considering the noise contained in generated data, we build a filter by using scores from verifiers and perplexity (PPL) from the generator. 
In detail, we select high-quality reasoning paths by setting a score threshold.
Moreover, we only keep the reasoning paths with no higher PPL than the ground truth solutions.
After filtering, we merge $D_{new}$ with $D$ and send it to Step 1.
Once the several iterations are completed, we obtain a powerful System 1\&2. More details can be found in~\cref{append:self_thinking}.

\subsection{Zero-shot Inference}
\label{sec:zero_shot_inference}

We simply perform cooperative inference as~\cref{fig:whole_model} Step 2 with trained System 1\&2 on unseen datasets.
After obtaining several reasoning paths with scores, we arrive at the final answer by weighted voting based on scores following~\citep{diverse-cot}.

\section{Experiments}
\label{sec:experiments}

\subsection{Experimental Setup}

\subsubsection{Datasets}

We consider several widely-used math word problem datasets: GSM8K~\cite{openai@gsm8k}, ASDiv-A~\citep{asdiv}, SingleOp~\citep{singleop}, SinlgeEq~\citep{singleeq} and MultiArith~\citep{multiarith}. (Details in~\cref{append:dataset_details}).
Following the general setting as in~\cite{DBLP:journals/corr/abs-2205-11916@zero-shot-reasoners,DBLP:journals/corr/abs-2201-11903@cot-ori}, we employ accuracy as the evaluation metric for all datasets.

\subsubsection{Baselines} 

For comparison under the zero-shot setting, the results of Instruct GPT-3 ($175$B) and PaLM ($540$B) with their various methods are from~\citet{DBLP:journals/corr/abs-2205-11916@zero-shot-reasoners}. 
The zero-shot$^*$ and zero-shot-CoT$^*$ imply not the standard prompt (see details in~\cref{append:baseline_settings}).
We also provide our generator as a baseline when compared to previous fine-tuning methods.
Regarding to sampling multiple solutions, we search $40$ paths with the same setting as Self-Consistency~\cite{DBLP:journals/corr/abs-2203-11171@self-consistency}.

For GSM8K, we select various powerful PLMs enhanced by the chain of thought prompt as baselines, including LaMDA ($137$B)~\cite{DBLP:journals/corr/abs-2201-08239@lamda}, GPT-3 ($175$B)~\cite{DBLP:conf/nips/BrownMRSKDNSSAA20@gpt3} and PaLM ($540$B)~\cite{DBLP:journals/corr/abs-2204-02311@palm}. 
Except for the few-shot methods, we also include a fine-tuned baseline that applies two GPT-3 ($175$B), one as the generator and the other as verifier~\cite{openai@gsm8k}.

\subsubsection{Implementation Details}

Since cooperative training requires a high-quality dataset with reasoning paths, we treat GSM8K~\cite{openai@gsm8k} as the seed dataset $D$ in~\cref{sec:cooperative_training}.
Unless otherwise, we employ GPT-J~\cite{gpt-j} as the generator and the step verifier, DeBERTa-large~\cite{DBLP:conf/iclr/HeLGC21@deberta} as the path verifier.
Since the default setting consists of two GPT-J ($6$B) and a DeBERTa-large ($0.4$B), we note our backbone as ``GPT-J $12$B'', which implies around 12.4 billion parameters in total.
During generation, we apply calculator as assistant following~\citet{openai@gsm8k}. We run all the experiments for 3 times and report the best result, detailed hyper-parameters setting can be found in~\cref{append:hyperparameters}.
Our zero-shot setting is similar to the transferring setting in T0~\cite{t0} and FLAN~\cite{flan}. 
All the training and testing procedures are done on a DGX station with $8$ A100 GPUs.

\begin{table}[tp]
\begin{center}
\begin{adjustbox}{max width=0.49\textwidth}
\begin{tabular}{ll|cc}
\toprule
Backbone           & Method                                   & SingleEq & MultiArith \\ \midrule
Instruct GPT-3 175B & zero-shot                                & 74.6     & 17.7       \\
                   & zero-shot$^*$                            & 78.7     & 22.7       \\
                   & zero-shot-CoT                            & 78.0     & 78.7       \\
                   & zero-shot-CoT$^*$                        & 78.7     & 79.3       \\
PaLM 540B          & zero-shot                                & -        & 25.5       \\
                   & zero-shot-CoT                            & -        & 66.1       \\
                   & $+$ Self-Consistency & -        & 89.0       \\ \midrule
GPT-J 12B       & CoRe (\textbf{ours})                            & \textbf{79.5}     & \textbf{97.5} \\
\bottomrule
\end{tabular}
\end{adjustbox}
\end{center}
\caption{
Zero-shot results on SingleEq and MultiArith. 
}
\label{table:zero_shot_results}
\end{table}

\begin{table*}[tp]
\begin{center}
\small
\begin{tabular}{ll|cccc}
\toprule
Backbone   & Method           & ASDiv-A         & SingleOp      & SingleEq      & MultiArith    \\ \midrule
{\ul Fine-tune}  & & & & & \\
&   Previous SoTA         & ~75.3$^{a}$    &  ~80.1$^{b}$   & ~72.3$^{c}$   & ~60.5$^{d}$ \\
 \midrule
{\ul Zero-shot}  &                     &               &               &               &                  \\

GPT-J 6B & Generator only & 51.7     & 53.2         & 49.2          & 77.3   \\
 & $+$ Self-Consistency & 63.7     & 59.6         & 60.2          & 92.3   \\
GPT-J 12B & CoRe \textbf{(ours)} & \textbf{90.5}     & \textbf{85.2}         & \textbf{79.5}          & \textbf{97.5}   \\
\bottomrule
\end{tabular}
\end{center}
\caption{
Zero-shot results v.s. previous fine-tuned SoTA results on math reasoning tasks. The previous SoTA baselines are obtained from:$a$: \citep{DBLP:conf/aaai/LanWZLD0ZL22@mwptoolkit}, $b$: LogicForm~\citep{DBLP:conf/ijcai/LiangHHLMS16@LogicForm-singleop-sota}, $c$: UNITDEP~\citep{DBLP:conf/aaai/RoyR17@UNITDEP-singleeq-sota}, $d$: Relevance and LCA operation classifier~\citep{multiarith}. The best scores are in \textbf{bold}.
}
\label{table:zero_shot_vs_finetuning}
\end{table*}

\subsection{Main Results}

\subsubsection{Zero-shot Results}
\label{sec:zero_shot_results}

\cref{table:zero_shot_results} presents main results on two mathematical reasoning datasets, demonstrating the zero-shot generalization ability.
CoRe achieves superior performance on both datasets, demonstrating its capability of mathematical reasoning on unseen datasets.
Note that the baselines are several dozen times larger than ours and still underperform our model.
The improvement might be explained by two potential reasons.
One is that applying the CoRe framework on PLMs can activate their reasoning ability, even though their scales are small ($\leq$ 100B).
Another one is that self-thinking can provide valuable self-produced data to teach Systems 1\&2. 
Therefore, the results present the effectiveness of cooperative working with System 1\&2 and self-thinking.

\subsubsection{Zero-shot v.s. Fine-tuning}
\label{sec:zero_shot_vs_finetuning}

We compare CoRe with previous fine-tuned SoTA baselines on four datasets, and results are presented in~\cref{table:zero_shot_vs_finetuning}.
To show the importance of cooperative reasoning, we apply our generator as a baseline.
The results demonstrate that without any guidance generator underperforms previous methods on most datasets. 
Despite the gain from self-consistency, it still lags behind other fine-tuned SoTAs. 
While after applying our method CoRe, it surpasses previous fine-tuned SoTAs on all datasets in a zero-shot setting. 
The results clearly demonstrate the capability of CoRe to greatly boost PLMs' reasoning ability.

\begin{table}[tp]
\begin{center}
\begin{adjustbox}{max width=0.49\textwidth}
\small
\begin{tabular}{ll|c}
\toprule
Backbone           & Method                                   & GSM8K \\ \midrule
{\ul few-shot}  & &  \\
LaMDA 137B   & CoT    & 17.1       \\
 & $+$ Self-Consistency & 27.7 \\
GPT-3 175B   & CoT    & 49.6     \\
            & $+$ Self-Consistency & - \\
PaLM 540B   & CoT    &  56.5    \\
            & $+$ Self-Consistency & \textbf{74.4} \\ \midrule
{\ul fine-tune}  & &  \\
GPT-3 350B    & -    & 57.0       \\
GPT-J 12B   & CoRe (\textbf{ours})     & {\ul 63.2} \\
\bottomrule
\end{tabular}
\end{adjustbox}
\end{center}
\caption{
Fine-tuning v.s. Few-shot results on GSM8K with various PLMs. Results are reported from~\citet{openai@gsm8k, DBLP:journals/corr/abs-2201-11903@cot-ori, DBLP:journals/corr/abs-2203-11171@self-consistency}. The best score is in \textbf{bold} and the second is {\ul underlined}.
}
\label{table:core_vs_cot_gsm8k}
\end{table}

\subsubsection{GSM8K Results}
\label{sec:fine_tuning_comparision}
Beyond improvements on zero-shot results, we observe that the fine-tuning setting can benefit a lot from our CoRe framework, as shown in~\cref{table:core_vs_cot_gsm8k}.
Compared to previous fine-tuned SoTA~\citep{openai@gsm8k} (GPT-3 350B), CoRe outperforms it with much fewer parameters, computation and inference time. 
Note that it samples $100$ solutions for each question while we only search $40$ paths.

For a comprehensive comparison, we include few-shot results with large-scale PLMs due to a limited number of ``fine-tune'' competitors.
With regard to few-shot methods applied on large-scale PLMs ($\geq$ 100B parameters), CoRe only underperforms PaLM-540B strengthened by chain of thought prompt and self-consistency, further proving the effectiveness of our method.

\begin{table}[tp]
\centering
\small
\begin{tabular}{lc|cc}
\toprule
Guidance        & $\alpha$ & SingleOp & MultiArith \\ \midrule
w/o verifiers    & -        & 59.6     & 92.3       \\
$V_{path}$ only    & 0        & 80.2     & 95.8       \\
$V_{path}$ + $V_{step}$ & 0.1      & 81.3     & 95.8       \\
                & 1        & \textbf{82.9}     & \textbf{96.8}       \\
\bottomrule
\end{tabular}
\caption{Zero-shot results with different levels of guidance from verifiers. $\alpha$ comes from~\cref{eq:score}.}
\label{table:abla_scores}
\end{table}

\begin{figure*}[!tp]
\centering
\includegraphics[width=0.98\textwidth]{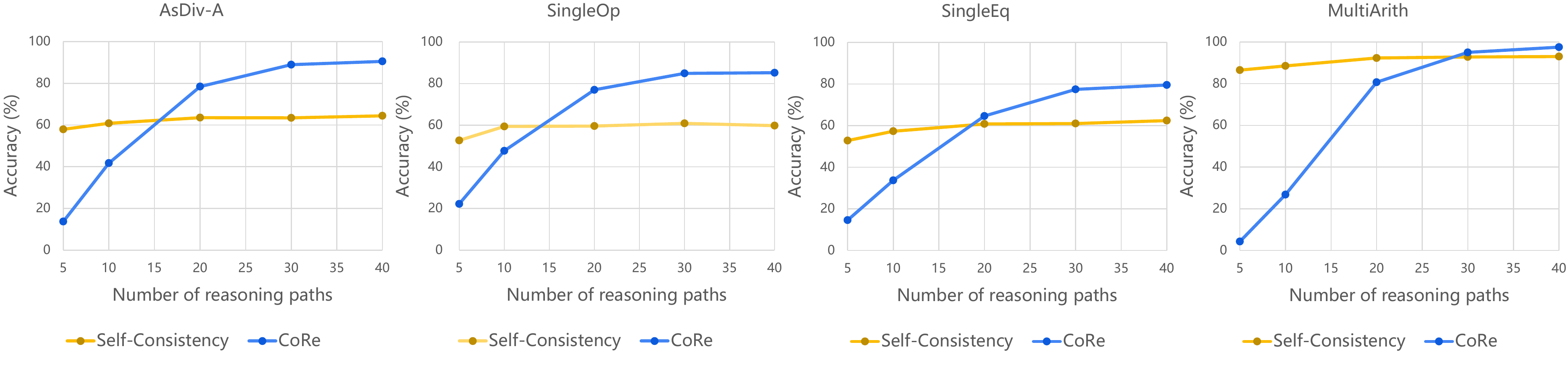}
\caption{
Zero-shot results with different search strategies in cooperative inference.
}
\label{fig:search_strategy}
\end{figure*}

\subsection{Ablation Study}
\label{sec:ablation_study}

\subsubsection{Is guidance important during path searching reasoning?}
\label{sec:abla_system12}

We argued that it is important to introduce guidance in the loop during reasoning path searching.
To validate this argument, we adjust the weight of reward provided by verifiers during reasoning. 
The experiments are conducted using models without self-thinking. 
\cref{table:abla_scores} summarizes the performance on zero-shot datasets with different settings of guidance.
For ``w/o verifiers'', the solutions are predicted by a generator only and applied with ``Self-Consistency''.
As demonstrated in~\cref{table:abla_scores}, guidance from $V_{path}$ can provide performance gains on SingleOp, with a $20.6\%$ absolute improvement. 
We further incorporate the guidance from the step-level verifier $V_{step}$. 
As described in~\cref{eq:score}, increasing the weight of reward ($\alpha$) from $V_{step}$, CoRe achieves a higher accuracy on both SingleOp and MultiArith.
Thanks to the feedback and guidance during the reasoning stage, the generator tends to explore more often on a path with a higher reward score. %
As a result, CoRe increases the accuracy on SingleOP from $59.6\%$ to $82.9\%$ and MultiArith from $92.3\%$ to $96.8\%$.

\subsubsection{How much does self-thinking boost the reasoning ability of a language model?}
\label{sec:abla_self_thinking_1}

To examine the effect of self-thinking, we explore it along with two axes: 1) the number of iterations and 2) the type of search strategy.
Since we apply the self-thinking procedure on the GSM8K dataset, we investigate the performance of models under different settings on GSM8K, as shown in~\cref{table:self_thinking_1}.
First, increasing the number of iterations can always improve the performance for both greedy decode and self-consistency. 
Our CoRe reaches saturation in one round, which might be attributed to the fact that System 1\&2 learns better and faster on self-generated data by collaborative working. 
Second, regardless of the search strategy, self-thinking consistently boost the model's performance, which verifies that self-thinking boost language model's reasoning ability.

\begin{table}[tp]
\centering
\small
\begin{tabular}{l|ccc}
\toprule
\# of iterations        & 0    & 1    & 2    \\ \midrule
Generator only (Greedy)         & 29.9 & 34.7 & \textbf{34.9} \\
Generator $+$ Self-Consistency & 42.0 & 43.1 & \textbf{45.9} \\
CoRe      & 60.0 & \textbf{63.2} & 61.6 \\ \bottomrule
\end{tabular}
\caption{Results on GSM8K with models undergone a different number of self-thinking iterations. Outcomes of various search strategies are provided.}
\label{table:self_thinking_1}
\end{table}

\subsubsection{Do self-thinking generalize to other datasets?}
\label{sec:abla_self_thinking_2}
We have performed self-thinking on GSM8K and proved that it improves the model's reasoning ability in~\ref{sec:abla_self_thinking_1}.
Furthermore, we explore whether the improvement on GSM8K comes at the cost of performance degradation on other datasets, i.e. the model overfits the dataset. 
As presented in~\cref{table:self_thinking_2}, we vary the number of self-thinking iterations for the generator and verifiers respectively and provide results on SingleOp and MultiArith. 
The results show that the performance of the generator suffers a little, but verifiers can eliminate this undesirable effect and benefit a lot from self-thinking. 
The best results are obtained when only the verifiers are further fine-tuned, with the $2.3\%$ and $0.7\%$ absolute improvement on the two datasets respectively. 
This observation implies that we can economize on training costs and time where target datasets are not included in self-thinking data.

\begin{table}[tp]
\small
\centering
\begin{tabular}{cc|cc}
\toprule
\multicolumn{2}{c|}{\# of iterations} & \multirow{2}{*}{SingleOp} & \multirow{2}{*}{MultiArith} \\ 
 Generator & Verifiers &               &               \\ \midrule
0         & 0         & 82.9          & 96.8          \\
0         & 1         & \textbf{85.2} & \textbf{97.5} \\
1         & 0         & 81.9          & 96.3          \\
1         & 1         & 83.3          & 97.2          \\
\bottomrule
\end{tabular}
\caption{Zero-shot results with a different number of self-thinking iterations for generator and verifiers respectively.}
\label{table:self_thinking_2}
\end{table}

\begin{table*}[!tp]
\setlength\tabcolsep{4pt}
\scriptsize
\centering

    \begin{tabu} to 1\textwidth { p{1.0cm} | p{13.3cm} | m{0.8cm}}
        \toprule

         \textbf{Question} & \textit{Sophia and Rose went together to the market to buy onions and potatoes. Rose bought 4 times the number of onions and potatoes Sophia bought. If Rose bought 12 onions and 4 potatoes, how many onions and potatoes in total did Sophia buy at the market?} & \textbf{Score}
         \\
         \midrule
         \textbf{Greedy Decode} & Rose bought 12 onions * \textcolor{red}{4 = 12 * 4 = 48 onions.} Rose bought 4 potatoes * \textcolor{red}{4 = 4 * 4 = 16 potatoes.} Sophia and Rose bought \textcolor{red}{48 onions + 16 potatoes = 64 onions and potatoes in total}. [ANS] \textcolor{red}{64}. & $-$
         \\
         \midrule
         \textbf{Most appropriate SC Path} & If Rose bought 12 onions at the market, then Sophia bought \textcolor{blue}{12 / 4 = 3 onions}. If Rose bought 4 potatoes at the market, then Sophia bought \textcolor{blue}{4 / 4 = 1 potato}. Since Rose bought 12 onions and 4 potatoes, Sophia bought \textcolor{blue}{3 + 1 = 4 onions and potatoes in total}. [ANS] \textcolor{blue}{4}. & $-$
         \\
         \midrule
         \textbf{Most consistent SC Path} & Rose bought \textcolor{red}{4 x 48 = 192 potatoes}. In total, Sophia bought \textcolor{red}{192 + 48 = 240 onions and potatoes} at the market. [ANS] \textcolor{red}{240}. & $-$
         \\
         \midrule
         \textbf{CoRe Path 1} & Since Rose bought 12 onions, this means there were \textcolor{blue}{12 / 4 = 3 onions that Sophia bought}. The number of potatoes that Sophia bought is 4, meaning there were \textcolor{blue}{4 / 4 = 1 potato that Sophia bought}. The total number of onions and potatoes that Sophia bought is \textcolor{blue}{3 + 1 = 4}. [ANS] \textcolor{blue}{4}. & $0.994$
         \\
         \midrule
         \textbf{CoRe Path 2} & The number of onions Rose bought is 12 onions / 4 = 3. Sophia bought 4 / 12 = 1 / 3 of the number of onions Rose bought. The number of potatoes Rose bought is 4 potatoes / 4 = 1\textcolor{red}{.33 of the number of potatoes Rose bought is 4 potatoes / 12 onions = 0.33}. The total number of onions and potatoes \textcolor{red}{Rose} bought at the market is 3 onions + 1 potatoes = 4. [ANS] 4.    & $0.363$
         \\
         \midrule
         \textbf{CoRe Path 3} & Let x be the number of onions Sophia bought. Rose bought \textcolor{red}{4 * x onions}. The total number of onions and potatoes Rose bought is \textcolor{red}{4 * x + x = 12. 5 * x = 12. x = 4}. Sophia bought \textcolor{red}{4 onions}. [ANS] 4. & $0.023$
         \\
         \bottomrule
    \end{tabu}
    \caption{GSM8K samples generated through different searching methods. SC refers to Self-Consistency. The erroneous parts of each path  are highlighted in \textcolor{red}{red} and the corresponding correct parts within other paths are \textcolor{blue}{blue}.}
    \label{table:case_study_core}
\end{table*}

\begin{table*}[!tp]
\setlength\tabcolsep{4pt}
\scriptsize
\centering
    \begin{tabu} to 1\textwidth { p{2.1cm} | p{13.0cm}}
        \toprule

         \textbf{Question} & \textit{Gunter is trying to count the jelly beans in a jar. He asks his friends how many they think are in the jar. One says 80. Another says 20 more than half the first one. A third says 25\% more than the first one. What is their average guess?} 
         \\
         \midrule
         \textbf{No Self-Thinking} & Half the first guess is 80 / 2 = 40. The third guess is \textcolor{red}{40 * .25 = 10.} Their total guesses are 80 \textcolor{red}{+ 20 + 40 + 10 = 150.} Their average guess is \textcolor{red}{150 / 4 = 37.5.} [ANS] \textcolor{red}{37.5}. 
         \\
         \midrule
         \textbf{With Self-Thinking} & Half the first friend's guess is 80 / 2 = 40. The second friend's guess is \textcolor{blue}{40 + 20 = 60.} The third friend's guess is \textcolor{blue}{80 * 1.25 = 100.} Their total guesses are \textcolor{blue}{80 + 60 + 100 = 240.} Their average guess is \textcolor{blue}{240 / 3 = 80.} [ANS] 80.
         \\
         \bottomrule
    \end{tabu}
    \caption{An example of GSM8K, model with self-thinking reasoned correctly, while the non-self-thinking model generated a wrong reasoning path and therefore failed.}
    \label{table:case_study_self_thinking}
\end{table*}

\subsubsection{How performance varies as the number of search iterations for different search strategies changes?}
\label{sec:abla_search_strategy}

As shown in~\cref{fig:search_strategy}, accuracy on $4$ datasets consistently increases along with the growth of search iterations for both search strategies. 
However, the scaling curves of self-consistency and CoRe are quite different.
The performance gain quickly saturates with self-consistency. Sampling $40$ paths can not further improve the accuracy, %
while the scaling curve of CoRe is much sharper. Due to the heuristic algorithm that requires the model to continue exploring on the previously generated paths, CoRe starts from a relatively lower level in the beginning, whereas the accuracy quickly improves as the number of search iterations increases. 
The result demonstrates the effectiveness of CoRe in searching reasoning paths, with a fast growth curve and a slow saturation rate.

\subsection{Case studies}
\label{sec:case_studies}

\subsubsection{Improvements from CoRe}

A typical exemplar from GSM8K is presented in~\cref{table:case_study_core}. 
Greedy decode fails to find a reasonable path due to the limited exploration in the output space. 
In contrast, self-consistency samples multiple reasoning paths randomly, resulting in a richer candidate set. 
Although it finds some right solutions occasionally, without any guidance, it fails to explore more frequently on the high-quality paths, thus ending up with a wrong answer obtained by majority voting as shown in the fourth row.

As a comparison, results generated by CoRe are listed with their scores. 
Similar to random sampling, the reasoning paths might be partially illogical, even though the final answers happen to be correct. 
Despite this challenge, CoRe is capable of distinguishing those poor-quality paths from the superior ones thanks to the verifiers. 
Adhering to the philosophy of cooperative reasoning we have emphasized, the verifiers managed to harness the generator throughout the reasoning procedure with the help of MCTS. 
Therefore, CoRe enjoys not only the advantage of having a diverse candidate set, but also the merit of being wiser and efficient during reasoning path searching.

\subsubsection{Improvements from Self-Thinking}

\cref{table:case_study_self_thinking} shows an example that the vanilla model failed to solve the given question, whereas after the self-thinking, the model rectified the faulty parts and successfully addressed it. 
This displays that self-thinking boosts language models' inner reasoning ability regardless of the search strategy, which is also proved in~\cref{sec:abla_self_thinking_1}.

\section{Discussion}
\label{sec:discussion}
Although we only fine-tune the language model on GSM8K due to the scarcity of QA datasets annotated with intermediate rationales, zero-shot results on several arithmetic datasets prove that basic reasoning capability is transferable across datasets within the same domain. 
This observation implies that when it comes to a new domain, we only need to collect a limited number of question-answer pairs with reasoning paths, model's reasoning ability can generalize to other unseen datasets and can be further strengthened by our approach CoRe according to the experimental results.

\section{Conclusions}
\label{sec:conclusion}
In this work, we mimic the dual system of human cognition to develop an effective reasoning framework for solving the math word problems. 
The proposed approach is consisting of two ingredients: the generator as System 1 and the verifiers as System 2, and overall reasoning is conducted based on their mutual reinforcement.
From the robustness and generalization aspects, CoRe activates superior reasoning ability of LMs, and thus outperforms PLMs that are dozens of times larger.

\section*{Limitations}
\label{sec:limitations}

The outcome on multiple datasets verifies the powerful reasoning ability, which even works on models with only several billion parameters.
However, our self-thinking procedure utilizes only one dataset, GSM8K, and the available training set size is only $7.5$K.
The main reason is the scarcity of high-quality datasets with rich reasoning paths.
And, collecting such data incurs huge computation costs and expensive human resources.
Another limitation is that we have not conducted experiments on bigger language models, such as GPT-3 and PaLM, due to the expensive usage costs and the fact of no open-source codes.
 In a nutshell, in the future, we will focus on collecting more high-quality labeled data and exploring our method on more powerful language models.

\section*{Ethics Statement}
\label{sec:ethical}

In this work, our CoRe shows impressive reasoning capability, however, it also comes with social risks.
Here, we summarize three possible ethical impacts: i) PLMs with bias, ii) generated data with social stereotypes and iii) problematic data environments.
Considering utilizing PLMs as backbones, several works present various potential risks in PLMs~\cite{lucy-bamman-2021-gender@bias-gpt3, DBLP:journals/corr/abs-2206-11993@lens-biases-gpt3}.
Fortunately, our method supports the replacement of different PLMs.
Therefore, we encourage deploying some risk-free PLMs, expecting to reduce the potential ethical risks.
Furthermore, once deploying harmful PLMs, the self-thinking process might generate several undesired data and those data are fed into language models, which deepens the bias and causes unintended social impacts.
For reducing the aforementioned cases, we suggest recording generated sentences.
In real-world applications, a good choice is to monitor generated content and then hand them over for human review.
In addition to the two risks posed by PLMs, the data in downstream tasks is of great concern.
In particular, private data might cause unpredictable influence because of their nature as a non-open source.
Therefore, we believe that a data cleaning workflow is necessary to mitigate potential risks, such as PrivateClean~\cite{DBLP:conf/sigmod/KrishnanWFGK16@private-clean}.
Finally, we encourage open debating about its utilization for increasing transparency and reducing the potential for misuse.

\section*{Acknowledgements}
\label{sec:acknowledgements}
This work was partly supported by the National Key Research and Development Program of China (No. 2020YFB1708200) ,  the "Graph Neural Network Project" of Ping An Technology (Shenzhen) Co., Ltd. and the Shenzhen Science and Technology Program (JCYJ20220818101001004).
{
\bibliography{anthology,custom}
\bibliographystyle{acl_natbib}
}

\clearpage
\appendix

\section{Dataset Details}
\label{append:dataset_details}

The mathematical reasoning datasets with details are as follows (Detailed description of the statistics in~\cref{table:append_datasets}).
We follow the licenses for their papers.

The dataset in fine-tuning:

\nbf{GSM8K~\cite{openai@gsm8k}} is a high-quality dataset with reasoning paths. 
It consists of 8.8K grade school math problems created by human writers, which are divided into a train set ($7.5$K) and a test set ($1.3$K).
The reasoning paths include $2$ to $8$ steps with considering basic arithmetic operations.
Furthermore, we conduct cooperative training and self-thinking on its training set.

The datasets in zero-shot inference:

\nbf{ASDiv-A~\cite{asdiv}} includes diverse math word problems, which are required to answer a number for each question.

\nbf{SingleOP~\cite{singleop}} is proposed with elementary math problems of a single operation.

\nbf{SingleEq~\cite{singleeq}} is construed with both single-step and multi-step math problems from mixed sources.

\nbf{MultiArith~\cite{multiarith}} includes elementary math problems with multiple steps.

\section{Experimental Settings}
\label{append:experimental_settings}

\subsection{Hyper-parameters Setting}
\label{append:hyperparameters}
For the generator and the step verifier, we train them for two epochs. The batch size is set to $16$. 
The learning rate~(LR) is set to $1e-5$ at the first epoch and $1e-6$ at the second epoch for generator.
On the hand of step verifier we apply the warmup method then linearly decaying scheduler, LR is set to $1e-6$ and warmup ratio is $0.1$. 

For the path verifier, we train it for three epochs with batch size set to $128$ and LR set to $1e-5$. 
Same LR scheduler as the step verifier has been applied for the path verifier. 
We set the gradient clip norm to $1.0$ and the sampling temperature to $0.7$.
The random seed is set to $19990303$ throughout the training process. 

For MCTS, we set max search iterations to $40$ during inference. In expansion, we search $20$ tokens each time. In order to avoid expanding too many homogeneous children for the same node, we simply penalize the probability of first token if it has appeared in other child nodes. We set the max token number to $300$ in roll out and limit the total token number of reasoning path to $400$.

\begin{table}[tp]
\centering
\small
\begin{tabular}{l|cc}
\toprule
Dataset    & \# of samples & Avg \# of words in questions \\ \midrule 
GSM8K      & 1319        & 46.9 
     \\
ASDiv-A    & 1218        & 29.2                        \\
SingleOp   & 562         & 20.9                        \\
SingleEq   & 508         & 27.2                        \\
MultiArith & 600         & 31.8                        \\
\bottomrule 
\end{tabular}
\caption{Dataset statistics.}
\label{table:append_datasets}
\end{table}

\subsection{Details of Training Verifiers}
\label{append:details_training_verifiers}
Before two verifiers are fine-tuned, we utilize the generator to sample 100 solutions for each question following ~\citet{openai@gsm8k}. 
Then we train the two verifiers on the generated data as described in \cref{sec:cooperative_training} Step 1.3.

\subsection{Details of Self-Thinking}
\label{append:self_thinking}
In each iteration of self-thinking, we initialize the model with the weights obtained from the previous round so as to save the computational costs. 
Since we use cooperative inference rather than random sampling to generate data for further training, solutions are expected more high-quality. 
Thus, the number of generated solutions $M$ mentioned in~\cref{sec:cooperative_training} is set to $50$ for saving computational cost and time. 
Due to the flexibility of MCTS, we have also tried to limit the time for searching rather than the number of iterations, which makes the total search time controllable and predictable. 
Moreover, this allows the model to adaptively adjust the final number of solutions searched for each question, due to the different levels of difficulty in questions.
In our experiments, we observe that setting the time limit to $320$ seconds provides better results than setting the iteration limit to $50$, while maintaining approximately the same time consumption. 
Therefore, we use time control to generate data during self-thinking.

\subsection{Baseline Settings}
\label{append:baseline_settings}

\begin{table*}[tp]
\centering
\small
\begin{tabular}{ll|cc}
\toprule
Backbone           & Method               & Reasoning Extraction Prompts & Answer Extraction Prompts       \\ \midrule
Instruct GPT-3 175B & zero-shot            & Let's think step by step.    & The answer (arabic numerals) is \\
                   & zero-shot$^*$        & Let's think step by step.    & The answer is                   \\
                   & zero-shot-CoT        & Let's think step by step.    & The answer (arabic numerals) is \\
                   & zero-shot-CoT$^*$    & Let's think step by step.    & The answer is                   \\ \midrule
PaLM 540B          & zero-shot            & Let's think step by step.    & The answer (arabic numerals) is \\
                   & zero-shot-CoT        & Let's think step by step.    & The answer (arabic numerals) is \\
                   & $+$ Self-Consistency & Let's think step by step.    & The answer (arabic numerals) is \\ \bottomrule
\end{tabular}
\caption{Prompt setting for few-shot baselines.}
\label{table:prompt_setting_baselines}
\end{table*}

As shown in~\cref{table:zero_shot_results}, the Instruct GPT-3 is based on text-davinci-002 version.
Moreover, since~\citet{DBLP:journals/corr/abs-2205-11916@zero-shot-reasoners} provides difference prompt setting, we list them in~\cref{table:prompt_setting_baselines}. 
For few-shot scenarios with the chain of thought prompts, we follow the original paper~\citep{DBLP:journals/corr/abs-2201-11903@cot-ori}.

\begin{table*}[tp]
\begin{center}
\small
\begin{tabular}{r|cccc}
\toprule
\# of reasoning paths     & ASDiv-A         & SingleOp      & SingleEq      & MultiArith    \\ \midrule 
{\ul Cobbe et al.} & & & &  \\

5 & 71.9 & 70.5 & 68.5 & 92.3   \\
10 & 76.9 & 73.1 & 74.6 & 95.0   \\
20 & 79.6 & 74.6 & 76.0 & 95.5  \\
30 & 81.4 & 76.2 & 76.2 & 95.2   \\
40 & 81.4 & 76.9 & 78.1 & 94.8   \\
\midrule
{\ul CoRe } & & & & \\
 5 & 13.7 & 22.2 & 14.6 & 4.3   \\
10 & 41.7 & 47.7 & 33.7 & 26.8   \\
20 & 78.4 & 77.0 & 64.6 & 80.7   \\
30 & 88.9 & 84.9 & 77.4 & 95.0 \\
40 & \textbf{90.5}     & \textbf{85.2}         & \textbf{79.5}          & \textbf{97.5}   \\
\bottomrule
\end{tabular}
\end{center}
\caption{
Comparison between \citet{openai@gsm8k} and CoRe with GPT-J as backbone model. The best scores are in \textbf{bold}.
}
\label{table:replicate_gsm8k}
\end{table*}
\section{Extended Experiments}
\label{append:extended_experiments}
This section we replicate the work of \citet{openai@gsm8k} with GPT-J and report the results in \cref{table:replicate_gsm8k} for comprehensive comparison. CoRe fully surpasses \citet{openai@gsm8k} when the number of reasoning paths reaches 30 and maintains a faster increasing rate after that. As a result, CoRe has a superior performance over \citet{openai@gsm8k} on all the datasets and achieves a $9.1\%$ and $8.3\%$ improvement compared to it on ASDiv-A and SingleOp.

\section{Future Work}
\label{append:future_work}

We focus on measuring our method in boosting the language model's arithmetic reasoning ability in this work. Nevertheless, we believe that our framework can also be applied to other reasoning tasks seamlessly, e.g., commonsense reasoning and symbolic reasoning. 
We choose arithmetic reasoning because it is the fundamental type of reasoning task.
Additionally, we believe solving arithmetic reasoning is the first step toward a general cognitive reasoning system. 
In the future, we will explore other reasoning tasks and put more effort into low-resource scenarios.

\end{document}